\documentclass{article} 

\usepackage[preprint]{neurips_2024}

\usepackage{hyperref}
\usepackage{url}
\usepackage[utf8]{inputenc} 
\usepackage[T1]{fontenc}    
\usepackage{hyperref}       
\usepackage{booktabs}       
\usepackage{algorithm}
\usepackage{algorithmic}
\usepackage{tikz}
\usetikzlibrary{arrows.meta,trees}

\usepackage{array}  
\usepackage{color}
\usepackage{newfloat}
\usepackage{listings}
\usepackage{booktabs}
\usepackage{float}
\usepackage{subfig}
\usepackage{tabularray}
\usepackage[most]{tcolorbox}
\usepackage{caption}
\usepackage{courier}
\usepackage{listings}
\usepackage{enumitem}
\usepackage{graphicx}
\usepackage{newfloat}
\usepackage{listings}
\usepackage{amssymb}   
\usepackage{amsmath}
\usepackage{tabularray}
\usepackage{xcolor}
\usepackage{makecell}
\usepackage{wrapfig}
\usepackage{multirow}

\title{DAMR: Efficient and Adaptive Context-Aware Knowledge Graph Question Answering with LLM-Guided MCTS}
\allowdisplaybreaks[4]
\newcommand{\method}{DAMR}

\author{%
  Yingxu Wang\\
  MBZUAI \\
  yingxv.wang@gmail.com
  \And
  Shiqi Fan\\
  PolyU\\
  fsq@mail.nwpu.edu.cn
  \And
  Mengzhu Wang\\
  Hebei University of Technology\\
  dreamkily@gmail.com
  \And
  Siyang Gao, Chao Wang\\
  CityU\\
  {siyangao@cityu.edu.hk, chanceycn@gmail.com}
  \And
  Nan Yin \\
  HKUST \\
  yinnan8911@gmail.com
}

\begin{document}

\maketitle
\begin{abstract}
Knowledge Graph Question Answering (KGQA) aims to interpret natural language queries and perform structured reasoning over knowledge graphs by leveraging their relational and semantic structures to retrieve accurate answers. 
Existing methods primarily follow either the retrieve-then-reason paradigm, which relies on Graph Neural Networks (GNNs) or heuristic rules to extract static candidate paths, or dynamic path generation strategies that employ large language models (LLMs) with prompting to jointly perform retrieval and reasoning. 
However, the former lacks adaptability due to static path extraction and the absence of contextual refinement, while the latter suffers from high computational costs and limited evaluation accuracy because of their dependence on fixed scoring functions and repeated LLM calls.
To address these issues, this paper proposes \textbf{D}ynamically \textbf{A}daptive \textbf{M}CTS-based \textbf{R}easoning (\method{}), a novel framework that integrates LLM-guided Monte Carlo Tree Search (MCTS) with adaptive path evaluation to enable efficient and context-aware KGQA. 
\method{} leverages MCTS as a backbone, where an LLM-based planner selects the top-$k$ semantically relevant relations at each expansion step to effectively reduce the search space. To enhance evaluation accuracy, we introduce a lightweight Transformer-based scorer that performs context-aware plausibility estimation by jointly encoding the question and relation sequence through cross-attention, thereby capturing fine-grained semantic shifts during multi-hop reasoning. Furthermore, to mitigate the scarcity of high-quality supervision, \method{} incorporates a dynamic pseudo-path refinement mechanism that periodically generates training signals from partial paths explored during search, enabling the scorer to continually adapt to the evolving distribution of reasoning trajectories.
Extensive experiments on multiple KGQA benchmarks show that \method{} significantly outperforms state-of-the-art methods. 
\end{abstract}

\section{Introduction}

Large Language Models (LLMs) have demonstrated impressive reasoning capabilities across diverse tasks, including mathematical problem solving~\citep{pei2025mathfusion,didolkar2024metacognitive}, commonsense inference~\citep{toroghi2024verifiable}, and open-domain question answering~\citep{zhao2023divknowqa}. 
Despite their generalization ability, LLMs often struggle in domain-specific scenarios due to the lack of grounded external knowledge, resulting in factual hallucinations and high inference costs~\citep{huang2025survey,wang2024factuality}. To address these limitations, recent efforts have explored integrating domain knowledge into LLM reasoning. A promising direction to overcome these limitations is Knowledge Graph Question Answering (KGQA)~\citep{dammu2025dynamic,saxena2020improving,choi2023nutrea}, which integrates relational structures into the reasoning process to provide factual grounding and structural interpretability. By combining the expressiveness of natural language with the precision of knowledge graphs, KGQA offers a scalable solution to improve factual consistency, reasoning transparency, and answer reliability~\citep{liu2025dual,yao2025learning}. 

Existing KGQA approaches can be broadly categorized into two paradigms according to how reasoning paths are constructed: retrieve-then-reason methods and dynamic path generation strategies. In the former, candidate paths are extracted prior to answer prediction, typically using Graph Neural Networks (GNNs) \citep{ma2025debate,yao2025learning} or rule-based heuristics \citep{fang2024karpa}. However, such methods exhibit limited adaptability, as GNNs fail to incorporate question-specific semantics during inference, while heuristic rules remain inherently inflexible and cannot support dynamic refinement of reasoning~\citep{liu2025dual,yao2025learning}. In contrast, dynamic path generation strategies unify retrieval and reasoning by constructing paths dynamically during question processing. These methods either leverage LLMs to iteratively generate paths through in-context learning or Chain-of-Thought (CoT)~\citep{sui2024fidelis,li2024framework}, or employ guided search techniques such as MCTS, where paths are incrementally expanded with a path scorer~\citep{ma2025deliberation,shen2025reasoning}. While offering greater flexibility, such approaches suffer from substantial computational overhead due to repeated LLM calls and limited evaluation accuracy, as static scorers cannot capture the evolving semantics of reasoning paths~\citep{chang2024survey,lee2024zero}.

This paper investigates the design of an adaptive KGQA framework to address challenges of computational inefficiency and limited path evaluation accuracy in dynamic reasoning. However, developing such a framework raises three key challenges: (1) \textit{How to modularize reasoning to reduce LLM overuse in search?} Inefficiency in dynamic KGQA stems from repeatedly invoking LLMs for relation retrieval and reasoning in multi-hop path construction~\citep{shen2025reasoning,long2025enhancing}. Although methods such as CoT and MCTS enable flexible exploration, they bind LLMs to every decision step, causing high inference cost and poor scalability. The challenge is to design a modular framework that leverages LLMs efficiently, guiding search without direct involvement at each step. (2) \textit{How to accurately evaluate evolving reasoning paths?} As multi-hop paths are incrementally constructed, their semantics evolve with each added relation and context. Yet existing methods rely on static scoring or shallow similarity metrics that fail to capture these semantic shifts~\citep{xu2024llm,sui2024fidelis}. This highlights the challenge of designing a path evaluation model that adaptively captures fine-grained changes conditioned on both the question and evolving relation sequence. (3) \textit{How to train a reliable evaluation model with limited supervision?} Accurate path ranking requires a well-calibrated scorer, yet dynamic reasoning generates many incomplete paths with only a few valid ones. This results in imbalanced, noisy supervision, especially for multi-hop questions where successful trajectories are scarce. While reinforcement learning has been explored~\citep{ma2024coevolving,zhai2024fine}, it suffers from sparse rewards and unstable optimization. The key challenge is to construct reliable learning signals from limited supervision to support adaptive scorer training.

To address these challenges, we propose \textbf{D}ynamically \textbf{A}daptive \textbf{M}CTS-based \textbf{R}easoning (\method{}), an efficient framework that integrates LLM-guided MCTS with context-aware semantic modeling for accurate and efficient KGQA.
\method{} employs an MCTS backbone, where an LLM-based planner dynamically guides path expansion by selecting semantically relevant relations at each step, thereby reducing the search space and improving answer identification.
For path evaluation, we introduce a lightweight Transformer-based scorer that jointly encodes the question and relation sequence via cross-attention, enabling context-sensitive plausibility estimation and capturing evolving semantics during multi-hop reasoning.
To mitigate supervision scarcity, \method{} further incorporates a dynamic pseudo-path mechanism that continuously adapts the scorer during search: partial paths from MCTS rollouts are ranked by predicted plausibility and converted into pseudo-path supervision pairs, amplifying signals from promising trajectories while suppressing noise from suboptimal ones.
Our contributions are summarized as follows:

\begin{itemize}[itemsep=2pt,topsep=0pt,parsep=0pt,leftmargin=*]
\item We study adaptive path reasoning in KGQA, where the key challenges lie in capturing the evolving semantics of multi-hop reasoning paths and ensuring computational efficiency during search, motivating the need for dynamic and context-aware reasoning strategies.
\item We propose \method{}, a novel framework that integrates MCTS with a dynamically adapted path evaluation model, enhancing evaluation accuracy while maintaining computational efficiency.
\item We conduct extensive experiments across multiple KGQA benchmarks, demonstrating that \method{} consistently outperforms state-of-the-art methods.
\end{itemize}
\section{Related work}

\paragraph{Knowledge Graph Question Answering (KGQA).} KGQA aims to enhance reasoning capabilities by incorporating external knowledge graphs to answer natural language questions~\citep{choi2023nutrea, xu2025memory}. Existing KGQA approaches can be broadly classified into two categories: retrieve-then-reason and dynamic path generation. The first category extracts candidate reasoning paths using Graph Neural Networks (GNNs)\citep{ma2025debate, yao2025learning} or rule-based heuristics\citep{fang2024karpa}, followed by LLM-based answer generation. While GNNs learn embeddings to identify relevant paths and rule-based methods apply predefined patterns~\citep{zhao2023divknowqa, liu2025dual}, these approaches lack the flexibility to adapt dynamically to question-specific context during inference. 
In contrast, dynamic path generation methods, such as CoT prompting~\citep{sui2024fidelis, li2024framework} and MCTS~\citep{ma2025deliberation, shen2025reasoning}, unify retrieval and reasoning for more flexible exploration. However, they suffer from high computational overhead due to repeated LLM calls, and static scorers often fail to adapt to evolving path semantics~\citep{long2025enhancing, luo2025kbqa,ma2024think}. To address these challenges, we propose an adaptive framework that integrates symbolic search with a fine-tuned evaluation model, aiming to improve both computational efficiency and reasoning accuracy in KGQA.

\noindent\textbf{Adaptive and Self-Improving Reasoning Models.} A promising approach to developing adaptive reasoning models is to frame the process within a reinforcement learning (RL) paradigm, where an agent learns a policy to navigate a state space. Early methods such as DeepPath~\citep{xiong2017deeppath} and MINERVA~\citep{das2018go} used RL to discover reasoning paths by rewarding the agent only when a correct answer is reached. However, this leads to the sparse rewards problem, as positive feedback arrives only after long action sequences, resulting in weak learning signals and poor exploration efficiency~\citep{zhai2024fine,chang2023learning}. To address this challenge, an alternative is self-training via pseudo-labeling, where the model learns from its own high-confidence predictions~\citep{lee2013pseudo,xie2020self}. While commonly used in semi-supervised learning, pseudo-labeling proves especially effective in reasoning tasks with limited supervision~\citep{wang2022semi,huang2025mllm}. Instead of relying on sparse terminal rewards, we leverage intermediate search paths as dynamic pseudo-paths, offering dense and adaptive supervision. This facilitates continual refinement of the path evaluator to better capture the evolving semantics of reasoning.
\section{Methodology}

\begin{wrapfigure}{r}{0.55\textwidth}
\vspace{-2.7cm}
\begin{minipage}{0.55\textwidth}
\centering
\includegraphics[scale=0.56]{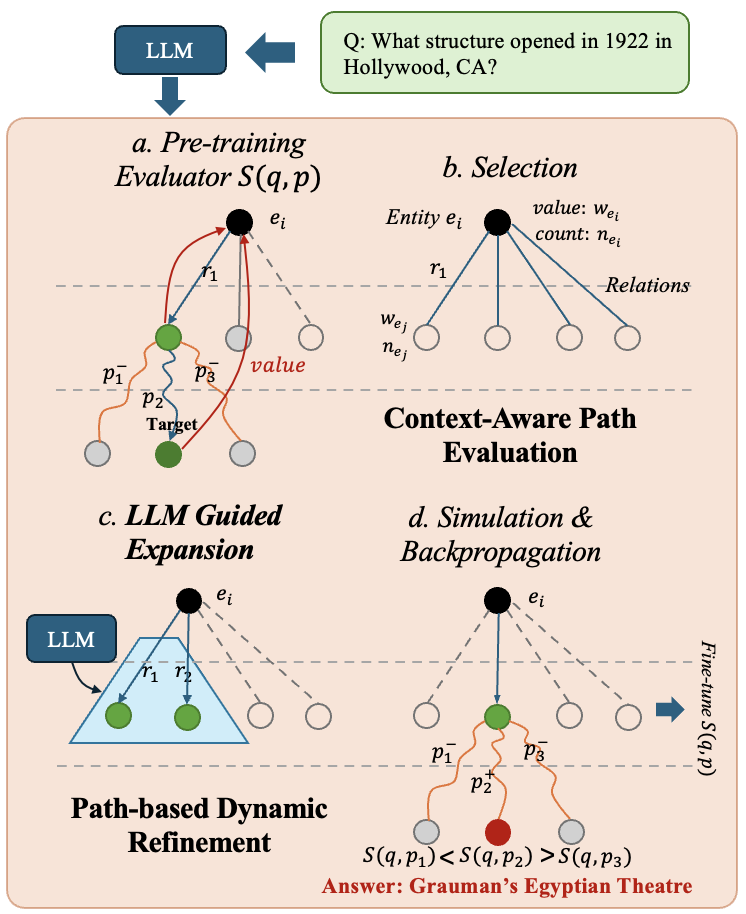}
\caption{Overview of \method{}. The reasoning process begins with an MCTS guided by an LLM-based planner, which selects top-$k$ semantically relevant relations at each expansion step. A context-aware path evaluator scores each candidate path during simulation. To enable continual adaptation, high-confidence pseudo-paths generated during search are used to dynamically fine-tune the evaluator.}
\label{framework}
\vspace{-1.4cm}
\end{minipage}
\end{wrapfigure}


\subsection{Problem Formulation}

We define Knowledge Graph Question Answering (KGQA) as the task of answering a natural language question by reasoning over a knowledge graph (KG). The KG is typically represented as a set of triples $\mathcal{K} = \{(e_s, r, e_o) \} \subseteq \mathcal{E} \times \mathcal{R} \times \mathcal{E}$, where $\mathcal{E}$ and $\mathcal{R}$ denote the sets of entities and binary relations.
The goal of KGQA is to find a set of answers $\mathcal{A}_q \subseteq \{(e_1,r_1,e_2), (e_2,r_2,e_3)\cdots\} $ for question $q$, such that a reasoning path through the KG leads from a topic entity to the correct answer. Formally, this is often framed as mapping $q$ to an executable query program $p_q$, where $\text{LLM}(p_q|{\mathcal{K}}) = \mathcal{A}_q$.

\subsection{Overview of Framework}
In this paper, we propose a dynamically adaptive reasoning framework \method{} for KGQA, as shown in Fig.~\ref{framework}. 
\method{} comprises three components: (1) \textbf{LLM Guided Expansion.} \method{} employs MCTS to incrementally expand reasoning paths, guided by an LLM-based planner that proposes relevant relations. This reduces computational overhead and enhances efficiency in KG exploration; (2) \textbf{Context-Aware Path Evaluation.} To capture the evolving semantics of reasoning paths, \method{} employs a lightweight Transformer-based scorer with cross-attention to jointly encode the question and path embeddings. This enables context-sensitive evaluation and improves the accuracy and relevance of multi-hop reasoning; (3) \textbf{Path-based Dynamic Refinement.} \method{} uses intermediate paths from MCTS as dynamic pseudo-paths to iteratively fine-tune the path evaluator, enhancing its ability to capture question-specific semantics and improving reasoning accuracy.

\subsection{LLM Guided Expansion}

A key challenge in KGQA is efficiently exploring the vast search space of multi-hop reasoning paths, especially under weak or no supervision. Existing methods often struggle to balance search efficiency and semantic relevance, resulting in either redundant exploration or missed correct paths~\citep{long2025enhancing,li2024simple}. To address this, the LLM-Guided Expansion module employs MCTS~\citep{kocsis2006bandit} as the backbone for symbolic path expansion. At each step, an LLM proposes semantically relevant relations, narrowing the search space and improving path quality, while MCTS ensures a balanced trade-off between exploration and exploitation.


Specifically, each node in the MCTS represents a reasoning state anchored at a specific entity in the KG. Given the current state, possible actions correspond to selecting an outgoing relation to extend the reasoning path. During the \textbf{Selection} phase, the search traverses from the root to a leaf node by recursively choosing children with the highest Upper Confidence Bound for Trees (UCT) score~\citep{kocsis2006bandit}, which is defined as:
\begin{equation}
\label{selection}
UCT = \frac{w_i}{n_i} + C \sqrt{\frac{\ln N}{n_i}},
\end{equation}
where $w_i$ denotes the accumulated reward of node $i$, $n_i$ its visit count, $N$ the visit count of its parent, and $C$ a constant that balances exploration and exploitation. This criterion guides the search to trade off between exploring new relations and exploiting promising paths.
In the \textbf{Expansion} phase, the selected leaf node is expanded by retrieving the outgoing relations of its entity $e_i$, denoted as $\mathcal{R}_{e_i} = \{r_1, r_2, \dots, r_n\}$. To avoid exhaustive branching, we prompt an LLM with the question $q$ and candidate relations $\mathcal{R}_{e_i}$, selecting the top-$k$ relations most semantically aligned with $q$:
\begin{equation}
\mathcal{R}_{top-k} = \text{LLM}(q, \mathcal{R}_{e_i}).
\end{equation}
These top-$k$ relations are then used to generate new child nodes, ensuring that expansion favors semantically meaningful directions and remains computationally efficient.


\subsection{Context-Aware Path Evaluation}

While LLM-guided expansion effectively narrows the search space by selecting semantically relevant relations, it does not ensure that all expanded paths remain valid in the broader reasoning context. As the search progresses, path semantics evolve dynamically, and early promising trajectories may later become misleading or irrelevant. Prior works attempt to filter or pre-plan paths to improve relevance~\citep{fang2024karpa,long2025eperm}, but still leave gaps in capturing evolving path semantics during search. To address this, we integrate a lightweight Transformer-based path scorer into MCTS’s simulation phase, employing cross-attention to jointly encode the question and current path, enabling adaptive evaluation that reflects evolving semantics.


\textbf{Context-Aware Path Evaluator.} 
In the \textbf{Simulation} phase, we assess the quality of candidate paths generated during MCTS rollouts. Given a question $q$ and a candidate relation path $p_r = (r_1, r_2, \ldots, r_l)$, where $p_r$ is formed by sequentially selecting relations during expansion, both $q$ and $p_r$ are first encoded using a pre-trained LLM. Let $\mathbf{z}_q \in \mathbb{R}^d$ denote the embedding of $q$ and $\mathbf{z}_{r_i} \in \mathbb{R}^d$ the embedding of relation $r_i$. To capture the sequential structure of $p_r$, we introduce a learnable positional encoding $\mathbf{e}_i^{\text{pos}}$ for each relation. The final input sequence is obtained by combining $\mathbf{z}_{r_i}$ with $\mathbf{e}_i^{\text{pos}}$ and feeding the sequence into a Transformer encoder:
\begin{equation}
\mathbf{E}_{p_r} = \text{Transformer}([\mathbf{z}_{r_1} + \mathbf{e}_1^{\text{pos}}, \ldots, \mathbf{z}_{r_l} + \mathbf{e}_l^{\text{pos}}]),
\end{equation}
where $\mathbf{e}_i^{\text{pos}} = \mathbf{E}^{\text{pos}}[i]$ is the positional embedding of the $i$-th hop, drawn from a trainable matrix $\mathbf{E}^{\text{pos}} \in \mathbb{R}^{L \times d}$, where $L$ is the maximum path length and $d$ the embedding dimension. To further incorporate question-specific information, we apply a cross-attention mechanism, allowing the encoded path representation $\mathbf{e}_{p_r}$ to attend to the question embedding $\mathbf{z}_q$:
\begin{equation}
\mathbf{H} = \mathbf{E}_{p_r} + \mathrm{CrossAttn}(\mathbf{E}_{p_r}, \mathbf{z}_q),    \;
    with\; \text{CrossAttn}(\mathbf{E}_{p_r}, \mathbf{z}_q)=\text{softmax}({\mathbf{E}_{p_r} \cdot \mathbf{z}_q^T}/{\sqrt{d_k}}) \cdot \mathbf{z}_q. 
\end{equation}
We then apply attention pooling over the relation representations $\mathbf{H}$ to obtain the hidden state of the relation path:
\begin{equation}
\mathbf{s}_{p_r} = \sum_{i=1}^l \alpha_i \textbf{h}_i, \quad \alpha = \text{Softmax}\big(\text{MLP}(\mathbf{H})\big),    
\end{equation}
where $\mathbf{h}_i$ is the hidden state of the $i$-th relation and $\alpha_i$ its learned attention weight. This pooling allows the model to emphasize informative steps along the reasoning path. The pooled path representation $\mathbf{s}_{p_r}$ is then concatenated with the question embedding $\mathbf{z}_q$, and the combined vector is passed through a multi-layer perceptron to compute the plausibility score of the question–path pair:
\begin{equation}
S(q, p_r) = \mathrm{MLP}\big([\mathbf{s}_{p_r}:\mathbf{z}_q]\big).    
\end{equation}
This context-aware evaluator assigns plausibility scores to partial reasoning paths by jointly encoding the question and relation sequence, providing accurate, context-sensitive guidance for MCTS.

\textbf{Pre-training of Evaluator.} To train the context-aware evaluator, we construct supervision by sampling positive and negative paths from local subgraphs. A path is positive if it links the head entity to a correct answer within a hop limit. Negatives come from two sources: hard negatives that terminate near but miss the answer, and random negatives from walks that avoid answer entities. Each training instance is a triplet $(q, p^+, p^-)$, with sequences zero-padded and masked for efficient batch training.

The model computes a plausibility score $S(q, p)$ for each question-path pair and is optimized using the Pair-wise Ranking loss~\citep{rendle2012bpr} to encourage higher scores for positive paths:
\begin{equation}
\label{pr}
    \mathcal{L}_{\text{PR}} = -\frac{1}{M} \sum_{i=1}^M \log \sigma\left( S(q, p^+_i) - S(q, p^-_i) \right),
\end{equation}
where $\sigma(\cdot)$ is the sigmoid function. This training strategy equips the evaluator with the ability to distinguish plausible reasoning paths, thereby improving the guidance signal during MCTS inference.



\subsection{Path-based Dynamic Refinement}

While LLM-guided expansion and semantic scoring improve path exploration, the static evaluator may fail to generalize to the evolving search space. To address this, we introduce a dynamic refinement mechanism that leverages high-confidence paths from MCTS rollouts as pseudo-paths. 
These pseudo-paths serve as supervision signals, enabling continual adaptation of the evaluator to new reasoning contexts without requiring additional labeled data.

Specifically, during \textbf{Backpropagation} phase, the plausibility score estimated by the context-aware path evaluator is propagated along the visited nodes in the MCTS tree after each simulation. For every entity $e_i$ on the simulated path, we update its visit count and aggregated value as follows:
\begin{equation}
    n_{e_i}= n_{e_i} + 1, \quad w_{e_i}= \frac{\sum_j n_{e_j}\cdot w_{e_j}}{\sum_j n_{e_j}},
\end{equation}
where $n_{e_i}$ is the visit count and $w_{e_i}$ is the aggregated value of entity $e_i$. The value is computed as a weighted average over its child nodes $\{e_j\}$, and reflects the plausibility scores $w_{e_j}$ assigned during simulation. These updates refine the UCT estimates used in future selection steps, progressively biasing the search toward high-quality reasoning paths.

To construct supervision signals for fine-tuning, we dynamically sample pseudo-path pairs ($\hat{p}'_i$, $\hat{p}'_j$) from the set of explored paths during MCTS. Instead of relying on the evaluator’s predictions, we assign pseudo-labels based on empirically grounded values derived from the search process. Specifically, for entity $e_i$ along a reasoning path $p_r$, we define its search value as:
$w_{e_i} = \frac{w_{p_r}}{n_{e_i}}$,
where $w_{p_r}$ is the cumulative reward from all rollouts passing through $p_r$, and $n_{e_i}$ is the visite count of entity $e_i$.
Given a pair of paths, we assign pseudo-labels based on their relative values:
\begin{equation}
    \left(\hat{p}^+,\, \hat{p}^-\right) =
\begin{cases}
(p'_i,\, p'_j), & \text{if } w_{e_i} > w_{e_j}, \\
(p'_j,\, p'_i), & \text{otherwise}.
\end{cases}
\end{equation}
The path evaluator is then fine-tuned using the Pair-wise Ranking loss in Eq.~\ref{pr}, encouraging higher scores for more promising paths.

\subsection{Reasoning Process}

The overall reasoning process is outlined in Appendix~\ref{sec:algorithm}. The framework first initializes a path evaluator to discriminate between plausible and implausible KG paths, providing a foundation for downstream search. During dynamic MCTS, the algorithm iteratively performs selection, expansion, simulation, and backpropagation. In expansion, an LLM-based planner adaptively selects the top-$k$ relations most relevant to the question, steering the search toward semantically meaningful paths. The evaluator guides simulation by prioritizing trajectories likely to yield correct answers, while pseudo-path pairs sampled during search are periodically used for refinement. Finally, entities reached by high-scoring paths are aggregated to form the answer set.

\section{Experiments}

\subsection{Experimental Settings}
\textbf{Datasets.} {To evaluate the effectiveness of \method{}, we conduct experiments on two widely used KGQA benchmarks: WebQSP~\cite{talmor2018web} and CWQ~\cite{yih2016value}. Following prior work~\cite{sun2023think,liu2025dual}, we uniformly sample 1,000 questions from the test sets of both datasets to evaluate the performance.} More details about datasets are provided in Appendix ~\ref{sec:dataset}.

\noindent\textbf{Baselines.} We compare \method{} with a comprehensive set of baselines. These baselines include: the semantic parsing methods, e.g., KV-Mem~\citep{miller2016key}, EmbedKGQA~\citep{saxena2020improving}, QGG~\citep{lan2020query}, NSM~\citep{he2021improving}, TransferNet~\citep{shi2021transfernet}, and KGT5~\citep{saxena2022sequence};
the retrieval-based methods, e.g., GraftNet~\citep{sun2018open}, PullNet~\citep{sun2019pullnet}, SR+NSM~\citep{zhang2022subgraph}, and SR+NSM+E2E~\citep{zhang2022subgraph};
the general LLMs, including 
Flan-T5-xl~\citep{chung2024scaling}, Alpaca-7B~\citep{taori2023stanford},
Llama3-8B~\citep{dubey2024llama}, Qwen2.5-7B~\citep{team2024qwen2}, ChatGPT~\citep{schulman2022chatgpt}, and ChatGPT+CoT~\citep{wei2022chain};
and recent LLMs with KG methods, including UniKGQA~\citep{jiang2022unikgqa},DECAF~\citep{yu2022decaf}, KD-CoT~\citep{wang2023knowledge},
Nutrea~\citep{choi2023nutrea}, ToG~\citep{sun2023think}, RoG~\citep{luo2023reasoning}, KAPING~\citep{baek2023knowledge}, ReasoningLM~\citep{jiang2023reasoninglm},
FiDeLis~\citep{sui2024fidelis}, GNN-RAG~\citep{mavromatis2024gnn},
DoG~\citep{ma2025debate},  DualR~\citep{liu2025dual}
, DP~\citep{ma2025deliberation}, and RwT~\citep{shen2025reasoning}. The details are provided in Appendix~\ref{sec:baseline}.

\begin{table*}[t]
\centering
\small
\begin{tabular}{clcccc}
\toprule
\multirow{2}{*}{\textbf{Type}} 
& \multirow{2}{*}{\textbf{Methods}} 
& \multicolumn{2}{c}{\textbf{WebQSP}} 
& \multicolumn{2}{c}{\textbf{CWQ}} \\
& & Hits@1 & F1 & Hits@1 & F1 \\
\midrule
\multirow{6}{*}{Semantic Parsing} 
& KV-Mem~\citep{miller2016key} & 46.7 & 34.5 & 18.4 & 15.7 \\
& EmbedKGQA~\citep{saxena2020improving} & 66.6 & - & 45.9 & - \\
& QGG~\citep{lan2020query} & 73.0 & 73.8 & 36.9 & 37.4 \\
& NSM~\citep{he2021improving} & 68.7 & 62.8 & 47.6 & 42.4 \\
& TransferNet~\citep{shi2021transfernet} & 71.4 & - & 48.6 & - \\
& KGT5~\citep{saxena2022sequence} & 56.1 & - & 36.5 & - \\
\midrule
\multirow{4}{*}{Retrieval} 
& GraftNet~\citep{sun2018open} & 66.4 & 60.4 & 36.8 & 32.7 \\
& PullNet~\citep{sun2019pullnet} & 68.1 & - & 45.9 & - \\
& SR+NSM~\citep{zhang2022subgraph} & 68.9 & 64.1 & 50.2 & 47.1 \\
& SR+NSM+E2E~\citep{zhang2022subgraph} & 69.5 & 64.1 & 49.3 & 46.3 \\
\midrule
\multirow{6}{*}{LLMs} 
& Flan-T5-xl~\cite{chung2024scaling} & 31.0 & - & 14.7 & - \\
& Alpaca-7B~\cite{taori2023stanford} & 51.8 & - & 27.4 & - \\
& Llama3-8B~\citep{dubey2024llama} & 30.3 & 25.7 & 30.5 & 27.8 \\
& Qwen2.5-7B~\citep{team2024qwen2} & 28.4 & 23.7 & 25.9 & 24.1 \\
& ChatGPT~\citep{schulman2022chatgpt} & 66.8 & - & 39.9 & - \\
& ChatGPT+CoT~\citep{wei2022chain} & 75.6 & - & 48.9 & - \\
\midrule
\multirow{13}{*}{LLMs+KGs} 
& UniKGQA~\citep{jiang2022unikgqa} & 77.2 & 72.2 & 51.2 & 49.0 \\
& DECAF~\cite{yu2022decaf} & 82.1 & 78.8 & 70.4 & - \\
& KD-CoT~\cite{wang2023knowledge} & 68.6 & 52.5 & 55.7 & - \\
& Nutrea~\citep{choi2023nutrea} & 77.4 & 72.7 & 53.6 & 49.5 \\
& ToG~\citep{sun2023think} & 81.9 & 76.0 & 68.5 & 60.2 \\
& RoG~\citep{luo2023reasoning} & 80.8 & 70.8 & 57.8 & 56.2 \\
& KAPING~\citep{baek2023knowledge} & 72.4 & 65.1 & 53.4 & 50.3 \\
& ReasoningLM~\citep{jiang2023reasoninglm} & 78.5 & 71.0 & 69.0 & 64.0 \\
& FiDeLis~\citep{sui2024fidelis} & 84.3 & 78.3 & 71.5 & 64.3 \\
& GNN-RAG~\citep{mavromatis2024gnn} & 82.8 & 73.5 & 62.8 & 60.4 \\
& DoG~\citep{ma2025debate} & 65.4 & 55.6 & 41.0 & 46.4 \\
& DualR~\citep{liu2025dual} & 81.5 & 71.6 & 65.3 & 62.1 \\
& DP~\citep{ma2025deliberation} & {87.5} & {81.4} & {75.8} & {69.4} \\
& RwT~\citep{shen2025reasoning} & 87.0 & 79.7 & 72.4 & 66.7 \\
\midrule
& \method{} & \textbf{94.0} & \textbf{81.7} & \textbf{78.0} & \textbf{75.1} \\
\bottomrule
\end{tabular}
\vspace{-2pt}
\caption{Performance comparison (\%) on WebQSP and CWQ datasets. \textbf{Bold} results indicate the best performance.}

\label{tab:main_results}
\end{table*}

\noindent\textbf{Implementation Details.} We implement the DAMR framework using PyTorch, and all experiments are conducted on NVIDIA A100 GPUs. The LLM-based planner is implemented with GPT-4.1~\citep{liu2023evaluating}, while question and relation embeddings are generated from Qwen3-Embedding-8B~\citep{yang2025qwen3} with an embedding dimension of 1024. For the path evaluation module, we use a 128-dimensional embedding and employ the Adam optimizer~\citep{kingma2014adam} with a learning rate of $1\times10^{-4}$ during pretraining and $1 \times 10^{-5}$ during fine-tuning. The model consists of two Transformer layers and is trained for 15 epochs in the pretraining stage and 10 epochs in the fine-tuning stage. Following~\citep{luo2023reasoning,yao2025learning,ma2025deliberation}, we evaluate \method{} using Hits@1 and F1 score, assessing answer correctness and overall accuracy for questions with potentially multiple correct answers.

\subsection{Performance Comparison}

We report the experimental results of \method{} in Table~\ref{tab:main_results}, benchmarking its performance against state-of-the-art baselines across KGQA datasets. From the results, we find that: 
(1) Semantic parsing and retrieval-based methods serve as early foundations for KGQA by extracting subgraphs and capturing structural semantics. However, embedding-based models struggle with complex relational patterns, while retrieval-based methods rely on rigid pipelines that limit generalization. In contrast, LLM with KG approaches combine the language understanding of LLMs with structured reasoning over KGs, enabling more flexible path exploration and improved adaptability to diverse, multi-hop queries.
(2) General-purpose LLMs, such as ChatGPT and Llama3-8B, show basic reasoning ability but often perform worse than methods that combine LLMs with KGs in KGQA tasks. This is mainly because they are not grounded in domain-specific knowledge, making them more likely to produce incorrect or made-up answers. 
(3) \method{} consistently outperforms all baselines across both datasets, showcasing its strong reasoning capability. This superior performance is driven by its integration of an LLM-based planner, which selectively retrieves relevant relations to reduce noise and guide the search toward high-quality reasoning paths, and a path evaluation model that is dynamically fine-tuned during search to capture semantic differences among candidate paths and accurately rank those most likely to yield correct answers.

\vspace{-1pt}
\subsection{Efficiency Analysis}
As shown in Table~\ref{tab:Efficiency}, \method{} achieves substantial improvements in computational efficiency. It reduces the average number of LLM calls to 7.1 on WebQSP and 16.8 on CWQ, with corresponding token usage of 3,931 and 9,266. These correspond to reductions of over 50\% in LLM calls and 75\% in token consumption relative to the strongest baseline. This efficiency is achieved by invoking the LLM only during the expansion phase of MCTS to select the top-$k$ semantically relevant relations, which effectively narrows the search space and avoids redundant reasoning steps that lead to unnecessary computational overhead. During simulation, the context-aware path evaluator efficiently assesses candidate paths based on question-path alignment without requiring any further LLM interaction or model inference. These design choices reduce both the frequency and verbosity of LLM usage while maintaining strong reasoning performance, making \method{} more efficient, scalable, and practically deployable than previous work.


\subsection{Ablation Study}\label{sec:ablation}

We conduct ablation studies to assess the contributions of key components in \method{}: (1) \method{} w/o PE: removing the path evaluation module; (2) \method{} w/o FT: disabling fine-tuning of the evaluator; and (3) \method{} w/ GPT-4.1: replacing the context-aware evaluator with a general LLM.

Experimental results are summarized in Table~\ref{tab:ablation_results}. From the results, we find that:
(1) Removing the path evaluation module (\method{} w/o PE) causes a noticeable performance drop on both datasets, underscoring its critical role in guiding the search process. Without this component, the model cannot effectively assess or rank candidate paths, leading to suboptimal reasoning and reduced accuracy.
(2) Compared to \method{} w/o FT, the proposed \method{} consistently achieves superior results on both datasets, highlighting the importance of the finetuning mechanism in the path evaluation module. This mechanism enables the model to adapt to the evolving distribution of explored paths, improving its ability to distinguish plausible from implausible reasoning trajectories.
(3) Replacing the context-aware path evaluation module with general LLMs yields degraded performance, confirming the advantage of our fine-tuned path scorer. By capturing fine-grained semantic distinctions among candidate paths, it provides more accurate evaluation signals, enhancing overall search effectiveness.











\begin{table}[t]
\centering
\begin{minipage}{\textwidth}
\begin{minipage}[t]{0.47\textwidth}
\makeatletter\def\@captype{table}
\centering
\footnotesize
\tabcolsep=3pt
\begin{tabular}{l@{\hspace{2em}}|cc|cc}
\toprule
\multirow{2}{*}{Method} & \multicolumn{2}{c|}{WebQSP} & \multicolumn{2}{c}{CWQ} \\
 & \#Tokens & \#Calls & \#Tokens & \#Calls \\
\midrule
DoG       & 22,538 & 30.9  & 37,741 & 58.1 \\
ToG       & 16,372 &  23.2 & 26,183 & 41.9 \\
RwT       & 10,680 & 15.1 & 17,885 &  28.6 \\
\midrule
\method{} &  \textbf{3,931} & \textbf{7.1} &  \textbf{9,266} &  \textbf{16.8} \\
\bottomrule
\end{tabular}
\vspace{0.5cm}
\caption{Statistics of average number of LLM calls and token consumption per question on WebQSP and CWQ datasets.}
\label{tab:Efficiency}
\end{minipage}
\hspace{0.18in}
\begin{minipage}[t]{0.48\textwidth}
\makeatletter\def\@captype{table}
\centering
\footnotesize
\tabcolsep=3pt
\begin{tabular}{l|cc|cc}
\toprule
\multirow{2}{*}{Method} & \multicolumn{2}{c}{WebQSP} & \multicolumn{2}{c}{CWQ} \\
 & Hits@1 & F1 & Hits@1 & F1 \\
\midrule

\method{} w/o PE & 91.2 & 78.2 & 74.3 & 72.1 \\

\method{} w/o FT & 91.9 & 80.1 & 75.1 & 73.0 \\

\method{} w/ GPT 4.1 & 92.5 & 79.8 & 74.9 & 72.4 \\




\midrule

\method{} & \textbf{94.0} & \textbf{81.7} & \textbf{78.0} & \textbf{75.1} \\

\bottomrule
\end{tabular}
\vspace{0.5cm}
\caption{The results of ablation studies on the WebQSP and CWQ datasets. \textbf{Bold} results indicate the best performance.}
\label{tab:ablation_results}
\end{minipage}
\end{minipage}
\vspace{-0.5cm}
\end{table}

\begin{wrapfigure}{r}{0.6\textwidth}
\vspace{-0.9cm}
\begin{minipage}[t]{0.6\textwidth}
        \centering
        \hspace{-1cm}
    \subfloat[Number of selected relations $k$]{\includegraphics[width=0.53\textwidth,height=0.29\textwidth]{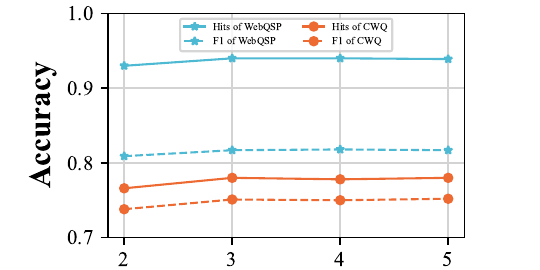}}
    \label{fig:mcts_top_k}
    \hspace{-0.5cm}
    \subfloat[Reasoning path length $L$]{\includegraphics[width=0.53\textwidth,height=0.29\textwidth]{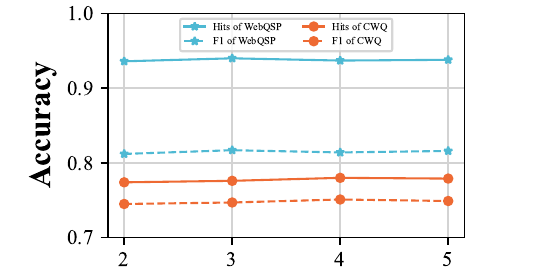}}
    \label{fig:mcts_length}
    \hspace{-1.3cm}
    \vspace{-0.2cm}
    \caption{Sensitivity analysis of hyperparameter on the WebQSP and CWQ datasets.}
        \label{fig:hyper_sensitive}
    \end{minipage}%
    \vspace{-0.4cm}
\end{wrapfigure}

\subsection{Sensitivity Analysis}\label{sec:sensitivity}

We conduct a sensitivity analysis to assess the impact of two key hyperparameters in \method{}: the number of selected relations $k$ and the maximum reasoning path length $L$. The parameter $k$ controls how many relations are proposed by the LLM-based planner at each step, while $L$ determines the number of reasoning hops allowed during path construction.

Figure \ref{fig:hyper_sensitive} illustrates how $k$ and  $L$ affect the performance of \method{} on the WebQSP and CWQ datasets. We vary $k$ and $L$ within the range of  $\{2, 3, 4, 5\}$. From the results, we observe that: (1) As shown in Figure~\ref{fig:hyper_sensitive}(a), increasing $k$ initially leads to performance gains, which then stabilize before experiencing a slight decline. While larger $k$ values encourage broader relational exploration, they may also introduce irrelevant candidates and increased computational cost. Conversely, smaller $k$ restrict the diversity of the search. To balance these trade-offs, we select a moderate $k=3$ as the default setting. (2) As shown in Figure~\ref{fig:hyper_sensitive}(b), on the WebQSP dataset, performance improves from $L=2$ to $3$, then fluctuates between $L=3$ and $5$, suggesting limited gains beyond three hops. In contrast, performance on the CWQ dataset steadily increases up to $L=4$ before slightly declining at $L=5$, reflecting its need for deeper reasoning due to more complex questions. Balancing effectiveness and efficiency across both datasets, we set $L=4$ as the default path length in all experiments. 
More results are provided in Appendix ~\ref{sec:experiments_results}.

\subsection{Impact of Different LLMs} 
\begin{wraptable}{r}{0.6\textwidth} 
\vspace{-29pt}
\footnotesize
\centering
\tabcolsep=6pt
\begin{tabular}{l|cc|cc}
\toprule
\multirow{2}{*}{Method} & \multicolumn{2}{c}{WebQSP} & \multicolumn{2}{c}{CWQ} \\
 & Hits@1 & F1 & Hits@1 & F1 \\
\midrule
\method{} (Llama2-13B) & 91.0 & 76.7 & 73.9 & 69.5 \\
\method{} (Qwen3-14B) & 91.5 & 77.8 & 74.4 & 70.1 \\
\method{} (GPT 4.1-mini) & 93.1 & 80.6 & 76.1 & 72.7 \\
\method{} (GPT 4.1) & \textbf{94.0} & \textbf{81.7} & \textbf{78.0} & \textbf{75.1} \\
\bottomrule
\end{tabular}
            \caption{Performance of \method{} using different LLM-based planners as backbones on different datasets.}\label{tab:backbone_results}
\end{wraptable}
To evaluate the impact of different LLM-based planners within the \method{} framework, we compare several backbones including Llama2-13B~\citep{roque2025evolution}, Qwen3-14B~\citep{team2024qwen2}, GPT 4.1 mini, and GPT 4.1, as shown in Table~\ref{tab:backbone_results}. Across both datasets, stronger LLMs consistently yield higher F1 and Hits scores, with GPT 4.1 achieving the best performance across all metrics. These results highlight the pivotal role of advanced LLMs in guiding relation selection and reasoning path expansion, where improved fluency, contextual awareness, and semantic precision translate into more accurate and faithful reasoning trajectories. Notably, the consistent performance gap between smaller and larger backbones illustrates the sensitivity of KGQA systems to the planner’s reasoning capability, reinforcing the necessity of leveraging high-capacity models when available. Overall, these findings emphasize the importance of backbone selection and further validate the design of \method{}, which capitalizes on powerful LLMs to achieve robust, generalizable, and effective multi-hop reasoning.

\begin{figure}[t]
    \centering
    \subfloat[One example from the WebQSP dataset.]{
    \label{fig:webqsp}    \includegraphics[width=0.47\linewidth]{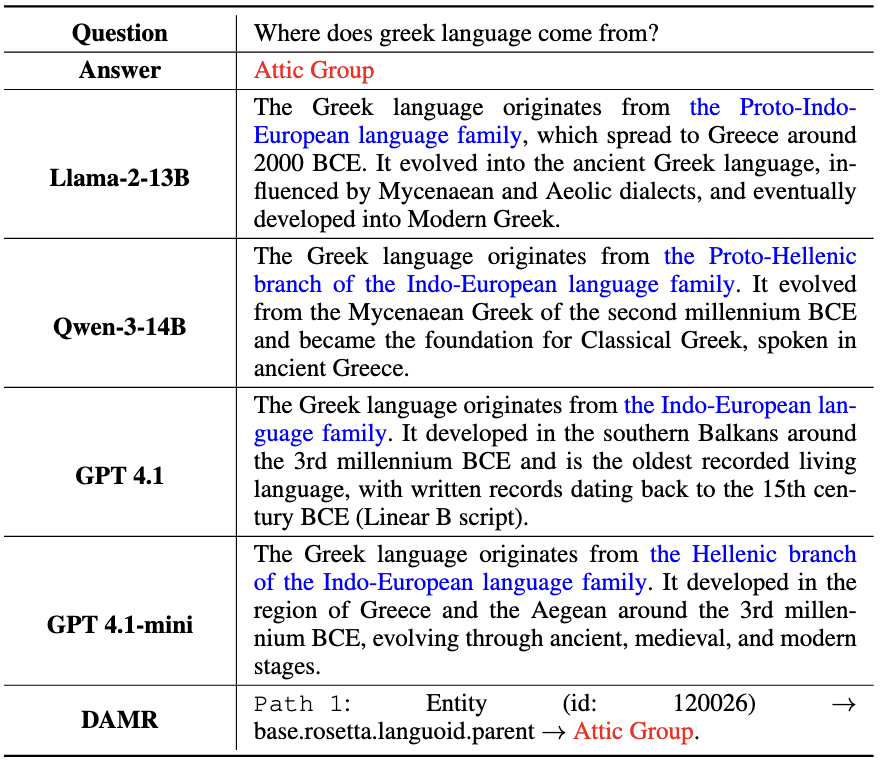}
    }
    \hspace{0.02\linewidth}
    \subfloat[One example from the CWQ dataset.]{
        \label{fig:cwq} \includegraphics[width=0.47\linewidth]{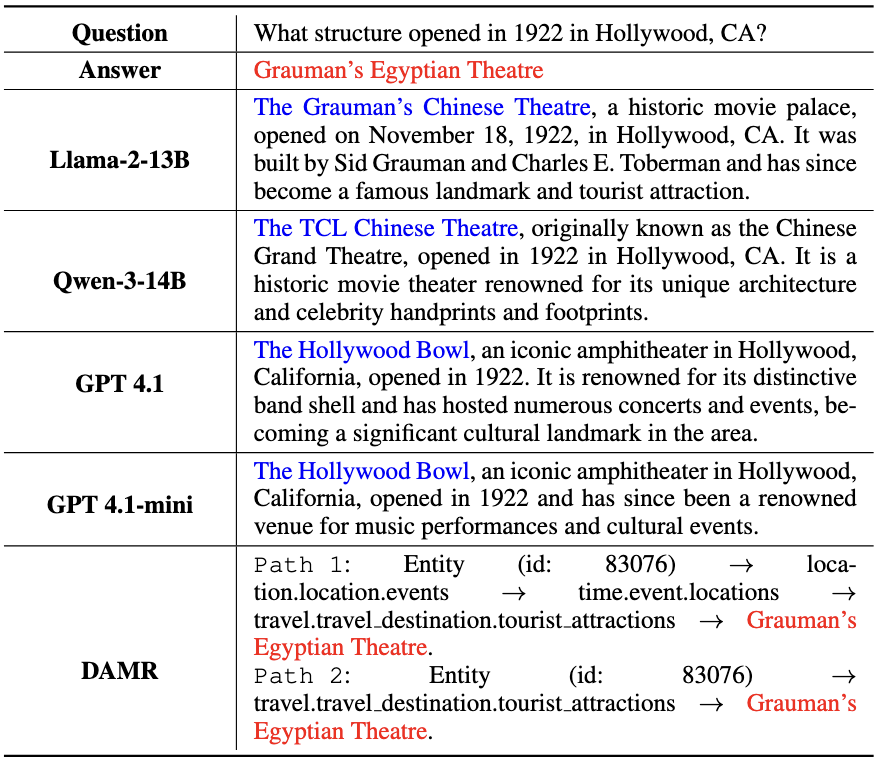}
    }
    \caption{Case studies of \method{} on the WebQSP and CWQ datasets. We highlight the correct answers in \textcolor{red}{Red} and the wrong answers in \textcolor{blue}{Blue}.}
    \label{fig:two_side_by_side}
\end{figure}

\subsection{Case study}

{Figure~\ref{fig:two_side_by_side} presents detailed case studies on the WebQSP and CWQ datasets comparing the reasoning process of \method{} with four representative LLMs: Llama-2-13B, Qwen-3-14B, GPT 4.1-mini, and GPT 4.1. In Figure~\ref{fig:webqsp}, the baseline LLMs generate fluent and seemingly plausible answers such as “Proto-Indo-European” or “Proto-Hellenic” when asked about the origin of the Greek language, yet fail to identify the correct answer \texttt{Attic Group}. In contrast, \method{} accurately predicts the correct entity by explicitly traversing the \textit{base.rosetta.languoid.parent} relation within the knowledge graph. In Figure~\ref{fig:cwq}, none of the baseline models are able to identify the structure that opened in Hollywood in 1922, while the proposed \method{} successfully locates \texttt{Grauman’s Egyptian Theatre} by explicitly following relevant relation paths in the knowledge graph and integrating complementary reasoning trajectories. These examples collectively highlight the strength of \method{} in grounding its reasoning in the knowledge graph and explicitly modeling multi-step reasoning paths, enabling it to provide accurate and faithful answers to complex, ontology-specific questions that general-purpose LLMs often struggle to resolve.}

\section{Conclusion}

In this work, we presented DAMR, a dynamically adaptive MCTS-based framework for knowledge graph question answering that integrates an LLM-guided planner, a context-aware path evaluator, and a dynamic pseudo-path refinement mechanism. By narrowing the search space, reducing redundant LLM calls, and continually adapting path evaluation, DAMR achieves efficient and accurate multi-hop reasoning. Extensive experiments on WebQSP and CWQ demonstrate that DAMR consistently surpasses state-of-the-art baselines in both performance and efficiency, while ablation and case studies highlight its interpretability and robustness. These results establish DAMR as a scalable solution for real-world KGQA and open avenues for extending adaptive symbolic–neural reasoning to broader domains such as scientific discovery and recommendation.


\bibliographystyle{plain}
\bibliography{main}

\begin{thebibliography}{10}

\bibitem{baek2023knowledge}
Jinheon Baek, Alham~Fikri Aji, and Amir Saffari.
\newblock Knowledge-augmented language model prompting for zero-shot knowledge graph question answering.
\newblock {\em arXiv preprint arXiv:2306.04136}, 2023.

\bibitem{chang2023learning}
Jonathan~D Chang, Kiante Brantley, Rajkumar Ramamurthy, Dipendra Misra, and Wen Sun.
\newblock Learning to generate better than your llm.
\newblock {\em arXiv preprint arXiv:2306.11816}, 2023.

\bibitem{chang2024survey}
Yupeng Chang, Xu~Wang, Jindong Wang, Yuan Wu, Linyi Yang, Kaijie Zhu, Hao Chen, Xiaoyuan Yi, Cunxiang Wang, Yidong Wang, et~al.
\newblock A survey on evaluation of large language models.
\newblock {\em ACM transactions on intelligent systems and technology}, 15(3):1--45, 2024.

\bibitem{choi2023nutrea}
Hyeong~Kyu Choi, Seunghun Lee, Jaewon Chu, and Hyunwoo~J Kim.
\newblock Nutrea: Neural tree search for context-guided multi-hop kgqa.
\newblock {\em Proceedings of the Conference on Neural Information Processing Systems}, 36:35954--35965, 2023.

\bibitem{chung2024scaling}
Hyung~Won Chung, Le~Hou, Shayne Longpre, Barret Zoph, Yi~Tay, William Fedus, Yunxuan Li, Xuezhi Wang, Mostafa Dehghani, Siddhartha Brahma, et~al.
\newblock Scaling instruction-finetuned language models.
\newblock {\em Journal of Machine Learning Research}, 25(70):1--53, 2024.

\bibitem{dammu2025dynamic}
Preetam Prabhu~Srikar Dammu, Himanshu Naidu, and Chirag Shah.
\newblock Dynamic-kgqa: A scalable framework for generating adaptive question answering datasets.
\newblock In {\em Proceedings of the 48th International ACM SIGIR Conference on Research and Development in Information Retrieval}, pages 3498--3508, 2025.

\bibitem{das2018go}
Rajarshi Das, Shehzaad Dhuliawala, Manzil Zaheer, Luke Vilnis, Ishan Durugkar, Akshay Krishnamurthy, Alex Smola, and Andrew McCallum.
\newblock Go for a walk and arrive at the answer: Reasoning over paths in knowledge bases using reinforcement learning.
\newblock In {\em Proceedings of the International Conference on Learning Representations}, 2018.

\bibitem{didolkar2024metacognitive}
Aniket Didolkar, Anirudh Goyal, Nan~Rosemary Ke, Siyuan Guo, Michal Valko, Timothy Lillicrap, Danilo Jimenez~Rezende, Yoshua Bengio, Michael~C Mozer, and Sanjeev Arora.
\newblock Metacognitive capabilities of llms: An exploration in mathematical problem solving.
\newblock {\em Proceedings of the Conference on Neural Information Processing Systems}, 37:19783--19812, 2024.

\bibitem{dubey2024llama}
Abhimanyu Dubey, Abhinav Jauhri, Abhinav Pandey, Abhishek Kadian, Ahmad Al-Dahle, Aiesha Letman, Akhil Mathur, Alan Schelten, Amy Yang, Angela Fan, et~al.
\newblock The llama 3 herd of models.
\newblock {\em arXiv e-prints}, pages arXiv--2407, 2024.

\bibitem{fang2024karpa}
Siyuan Fang, Kaijing Ma, Tianyu Zheng, Xinrun Du, Ningxuan Lu, Ge~Zhang, and Qingkun Tang.
\newblock Karpa: A training-free method of adapting knowledge graph as references for large language model's reasoning path aggregation.
\newblock {\em arXiv preprint arXiv:2412.20995}, 2024.

\bibitem{he2021improving}
Gaole He, Yunshi Lan, Jing Jiang, Wayne~Xin Zhao, and Ji-Rong Wen.
\newblock Improving multi-hop knowledge base question answering by learning intermediate supervision signals.
\newblock In {\em Proceedings of the International ACM Conference on Web Search \& Data Mining}, pages 553--561, 2021.

\bibitem{huang2025mllm}
Jiaxin Huang, Runnan Chen, Ziwen Li, Zhengqing Gao, Xiao He, Yandong Guo, Mingming Gong, and Tongliang Liu.
\newblock Mllm-for3d: Adapting multimodal large language model for 3d reasoning segmentation.
\newblock {\em arXiv preprint arXiv:2503.18135}, 2025.

\bibitem{huang2025survey}
Lei Huang, Weijiang Yu, Weitao Ma, Weihong Zhong, Zhangyin Feng, Haotian Wang, Qianglong Chen, Weihua Peng, Xiaocheng Feng, Bing Qin, et~al.
\newblock A survey on hallucination in large language models: Principles, taxonomy, challenges, and open questions.
\newblock {\em ACM Transactions on Information Systems}, 43(2):1--55, 2025.

\bibitem{jiang2023reasoninglm}
Jinhao Jiang, Kun Zhou, Wayne~Xin Zhao, Yaliang Li, and Ji-Rong Wen.
\newblock Reasoninglm: Enabling structural subgraph reasoning in pre-trained language models for question answering over knowledge graph.
\newblock {\em arXiv preprint arXiv:2401.00158}, 2023.

\bibitem{jiang2022unikgqa}
Jinhao Jiang, Kun Zhou, Wayne~Xin Zhao, and Ji-Rong Wen.
\newblock Unikgqa: Unified retrieval and reasoning for solving multi-hop question answering over knowledge graph.
\newblock {\em arXiv preprint arXiv:2212.00959}, 2022.

\bibitem{kingma2014adam}
Diederik~P Kingma.
\newblock Adam: A method for stochastic optimization.
\newblock {\em arXiv preprint arXiv:1412.6980}, 2014.

\bibitem{kocsis2006bandit}
Levente Kocsis and Csaba Szepesv{\'a}ri.
\newblock Bandit based monte-carlo planning.
\newblock In {\em European conference on machine learning}, pages 282--293. Springer, 2006.

\bibitem{lan2020query}
Yunshi Lan and Jing Jiang.
\newblock Query graph generation for answering multi-hop complex questions from knowledge bases.
\newblock Proceedings of the Annual Meeting of the Association for Computational Linguistics, 2020.

\bibitem{lee2013pseudo}
Dong-Hyun Lee et~al.
\newblock Pseudo-label: The simple and efficient semi-supervised learning method for deep neural networks.
\newblock In {\em Workshop on challenges in representation learning, ICML}, number~2, page 896. Atlanta, 2013.

\bibitem{lee2024zero}
Seongmin Lee, Jaewook Shin, Youngjin Ahn, Seokin Seo, Ohjoon Kwon, and Kee-Eung Kim.
\newblock Zero-shot multi-hop question answering via monte-carlo tree search with large language models.
\newblock {\em arXiv preprint arXiv:2409.19382}, 2024.

\bibitem{li2024simple}
Mufei Li, Siqi Miao, and Pan Li.
\newblock Simple is effective: The roles of graphs and large language models in knowledge-graph-based retrieval-augmented generation.
\newblock {\em arXiv preprint arXiv:2410.20724}, 2024.

\bibitem{li2024framework}
Yading Li, Dandan Song, Changzhi Zhou, Yuhang Tian, Hao Wang, Ziyi Yang, and Shuhao Zhang.
\newblock A framework of knowledge graph-enhanced large language model based on question decomposition and atomic retrieval.
\newblock In {\em Proceedings of the Conference on Empirical Methods in Natural Language Processing}, pages 11472--11485, 2024.

\bibitem{liu2025dual}
Guangyi Liu, Yongqi Zhang, Yong Li, and Quanming Yao.
\newblock Dual reasoning: A gnn-llm collaborative framework for knowledge graph question answering.
\newblock In {\em The Second Conference on Parsimony and Learning (Proceedings Track)}, 2025.

\bibitem{liu2023evaluating}
Hanmeng Liu, Ruoxi Ning, Zhiyang Teng, Jian Liu, Qiji Zhou, and Yue Zhang.
\newblock Evaluating the logical reasoning ability of chatgpt and gpt-4.
\newblock {\em arXiv preprint arXiv:2304.03439}, 2023.

\bibitem{long2025eperm}
Xiao Long, Liansheng Zhuang, Aodi Li, Minghong Yao, and Shafei Wang.
\newblock Eperm: An evidence path enhanced reasoning model for knowledge graph question and answering.
\newblock In {\em Proceedings of the AAAI Conference on Artificial Intelligence}, volume~39, pages 12282--12290, 2025.

\bibitem{long2025enhancing}
Xiao Long, Liansheng Zhuang, Chen Shen, Shaotian Yan, Yifei Li, and Shafei Wang.
\newblock Enhancing large language models with reward-guided tree search for knowledge graph question and answering.
\newblock {\em arXiv preprint arXiv:2505.12476}, 2025.

\bibitem{luo2025kbqa}
Haoran Luo, Yikai Guo, Qika Lin, Xiaobao Wu, Xinyu Mu, Wenhao Liu, Meina Song, Yifan Zhu, Luu~Anh Tuan, et~al.
\newblock Kbqa-o1: Agentic knowledge base question answering with monte carlo tree search.
\newblock {\em arXiv preprint arXiv:2501.18922}, 2025.

\bibitem{luo2023reasoning}
Linhao Luo, Yuan-Fang Li, Gholamreza Haffari, and Shirui Pan.
\newblock Reasoning on graphs: Faithful and interpretable large language model reasoning.
\newblock {\em arXiv preprint arXiv:2310.01061}, 2023.

\bibitem{ma2024coevolving}
Hao Ma, Tianyi Hu, Zhiqiang Pu, Liu Boyin, Xiaolin Ai, Yanyan Liang, and Min Chen.
\newblock Coevolving with the other you: Fine-tuning llm with sequential cooperative multi-agent reinforcement learning.
\newblock {\em Proceedings of the Conference on Neural Information Processing Systems}, pages 15497--15525, 2024.

\bibitem{ma2025debate}
Jie Ma, Zhitao Gao, Qi~Chai, Wangchun Sun, Pinghui Wang, Hongbin Pei, Jing Tao, Lingyun Song, Jun Liu, Chen Zhang, et~al.
\newblock Debate on graph: a flexible and reliable reasoning framework for large language models.
\newblock In {\em Proceedings of the AAAI Conference on Artificial Intelligence}, number~23, pages 24768--24776, 2025.

\bibitem{ma2025deliberation}
Jie Ma, Ning Qu, Zhitao Gao, Rui Xing, Jun Liu, Hongbin Pei, Jiang Xie, Linyun Song, Pinghui Wang, Jing Tao, et~al.
\newblock Deliberation on priors: Trustworthy reasoning of large language models on knowledge graphs.
\newblock {\em arXiv preprint arXiv:2505.15210}, 2025.

\bibitem{ma2024think}
Shengjie Ma, Chengjin Xu, Xuhui Jiang, Muzhi Li, Huaren Qu, Cehao Yang, Jiaxin Mao, and Jian Guo.
\newblock Think-on-graph 2.0: Deep and faithful large language model reasoning with knowledge-guided retrieval augmented generation.
\newblock {\em arXiv preprint arXiv:2407.10805}, 2024.

\bibitem{mavromatis2024gnn}
Costas Mavromatis and George Karypis.
\newblock Gnn-rag: Graph neural retrieval for large language model reasoning.
\newblock {\em arXiv preprint arXiv:2405.20139}, 2024.

\bibitem{miller2016key}
Alexander Miller, Adam Fisch, Jesse Dodge, Amir-Hossein Karimi, Antoine Bordes, and Jason Weston.
\newblock Key-value memory networks for directly reading documents.
\newblock {\em arXiv preprint arXiv:1606.03126}, 2016.

\bibitem{pei2025mathfusion}
Qizhi Pei, Lijun Wu, Zhuoshi Pan, Yu~Li, Honglin Lin, Chenlin Ming, Xin Gao, Conghui He, and Rui Yan.
\newblock Mathfusion: Enhancing mathematical problem-solving of llm through instruction fusion.
\newblock {\em arXiv preprint arXiv:2503.16212}, 2025.

\bibitem{rendle2012bpr}
Steffen Rendle, Christoph Freudenthaler, Zeno Gantner, and Lars Schmidt-Thieme.
\newblock Bpr: Bayesian personalized ranking from implicit feedback.
\newblock {\em arXiv preprint arXiv:1205.2618}, 2012.

\bibitem{roque2025evolution}
Lu{\'\i}s Roque.
\newblock The evolution of llama: From llama 1 to llama 3.1, 2025.

\bibitem{saxena2022sequence}
Apoorv Saxena, Adrian Kochsiek, and Rainer Gemulla.
\newblock Sequence-to-sequence knowledge graph completion and question answering.
\newblock {\em arXiv preprint arXiv:2203.10321}, 2022.

\bibitem{saxena2020improving}
Apoorv Saxena, Aditay Tripathi, and Partha Talukdar.
\newblock Improving multi-hop question answering over knowledge graphs using knowledge base embeddings.
\newblock In {\em Proceedings of the 58th annual meeting of the association for computational linguistics}, pages 4498--4507, 2020.

\bibitem{schulman2022chatgpt}
John Schulman, Barret Zoph, Christina Kim, Jacob Hilton, Jacob Menick, Jiayi Weng, Juan Felipe~Ceron Uribe, Liam Fedus, Luke Metz, Michael Pokorny, et~al.
\newblock Chatgpt: Optimizing language models for dialogue.
\newblock {\em OpenAI blog}, 2(4), 2022.

\bibitem{shen2025reasoning}
Tiesunlong Shen, Jin Wang, Xuejie Zhang, and Erik Cambria.
\newblock Reasoning with trees: Faithful question answering over knowledge graph.
\newblock In {\em Proceedings of the Annual Meeting of the Association for Computational Linguistics}, pages 3138--3157, 2025.

\bibitem{shi2021transfernet}
Jiaxin Shi, Shulin Cao, Lei Hou, Juanzi Li, and Hanwang Zhang.
\newblock Transfernet: An effective and transparent framework for multi-hop question answering over relation graph.
\newblock {\em arXiv preprint arXiv:2104.07302}, 2021.

\bibitem{sui2024fidelis}
Yuan Sui, Yufei He, Nian Liu, Xiaoxin He, Kun Wang, and Bryan Hooi.
\newblock Fidelis: Faithful reasoning in large language model for knowledge graph question answering.
\newblock {\em arXiv preprint arXiv:2405.13873}, 2024.

\bibitem{sun2019pullnet}
Haitian Sun, Tania Bedrax-Weiss, and William~W Cohen.
\newblock Pullnet: Open domain question answering with iterative retrieval on knowledge bases and text.
\newblock {\em arXiv preprint arXiv:1904.09537}, 2019.

\bibitem{sun2018open}
Haitian Sun, Bhuwan Dhingra, Manzil Zaheer, Kathryn Mazaitis, Ruslan Salakhutdinov, and William~W Cohen.
\newblock Open domain question answering using early fusion of knowledge bases and text.
\newblock {\em arXiv preprint arXiv:1809.00782}, 2018.

\bibitem{sun2023think}
Jiashuo Sun, Chengjin Xu, Lumingyuan Tang, Saizhuo Wang, Chen Lin, Yeyun Gong, Lionel~M Ni, Heung-Yeung Shum, and Jian Guo.
\newblock Think-on-graph: Deep and responsible reasoning of large language model on knowledge graph.
\newblock {\em arXiv preprint arXiv:2307.07697}, 2023.

\bibitem{talmor2018web}
Alon Talmor and Jonathan Berant.
\newblock The web as a knowledge-base for answering complex questions.
\newblock {\em arXiv preprint arXiv:1803.06643}, 2018.

\bibitem{taori2023stanford}
Rohan Taori, Ishaan Gulrajani, Tianyi Zhang, Yann Dubois, Xuechen Li, Carlos Guestrin, Percy Liang, and Tatsunori~B Hashimoto.
\newblock Stanford alpaca: An instruction-following llama model, 2023.

\bibitem{team2024qwen2}
Qwen Team.
\newblock Qwen2 technical report.
\newblock {\em arXiv preprint arXiv:2407.10671}, 2024.

\bibitem{toroghi2024verifiable}
Armin Toroghi, Willis Guo, Ali Pesaranghader, and Scott Sanner.
\newblock Verifiable, debuggable, and repairable commonsense logical reasoning via llm-based theory resolution.
\newblock In {\em Proceedings of the Conference on Empirical Methods in Natural Language Processing}, 2024.

\bibitem{wang2023knowledge}
Keheng Wang, Feiyu Duan, Sirui Wang, Peiguang Li, Yunsen Xian, Chuantao Yin, Wenge Rong, and Zhang Xiong.
\newblock Knowledge-driven cot: Exploring faithful reasoning in llms for knowledge-intensive question answering.
\newblock {\em arXiv preprint arXiv:2308.13259}, 2023.

\bibitem{wang2022semi}
Yuchao Wang, Haochen Wang, Yujun Shen, Jingjing Fei, Wei Li, Guoqiang Jin, Liwei Wu, Rui Zhao, and Xinyi Le.
\newblock Semi-supervised semantic segmentation using unreliable pseudo-labels.
\newblock In {\em The IEEE/CVF Conference on Computer Vision and Pattern Recognition}, pages 4248--4257, 2022.

\bibitem{wang2024factuality}
Yuxia Wang, Minghan Wang, Muhammad~Arslan Manzoor, Fei Liu, Georgi Georgiev, Rocktim~Jyoti Das, and Preslav Nakov.
\newblock Factuality of large language models: A survey.
\newblock {\em arXiv preprint arXiv:2402.02420}, 2024.

\bibitem{wei2022chain}
Jason Wei, Xuezhi Wang, Dale Schuurmans, Maarten Bosma, Fei Xia, Ed~Chi, Quoc~V Le, Denny Zhou, et~al.
\newblock Chain-of-thought prompting elicits reasoning in large language models.
\newblock {\em Proceedings of the Conference on Neural Information Processing Systems}, 35:24824--24837, 2022.

\bibitem{xie2020self}
Qizhe Xie, Minh-Thang Luong, Eduard Hovy, and Quoc~V Le.
\newblock Self-training with noisy student improves imagenet classification.
\newblock In {\em The IEEE/CVF Conference on Computer Vision and Pattern Recognition}, pages 10687--10698, 2020.

\bibitem{xiong2017deeppath}
Wenhan Xiong, Thien Hoang, and William~Yang Wang.
\newblock Deeppath: A reinforcement learning method for knowledge graph reasoning.
\newblock In {\em Proceedings of the Conference on Empirical Methods in Natural Language Processing}, pages 564--573, 2017.

\bibitem{xu2024llm}
Mufan Xu, Kehai Chen, Xuefeng Bai, Muyun Yang, Tiejun Zhao, and Min Zhang.
\newblock Llm-based discriminative reasoning for knowledge graph question answering.
\newblock {\em arXiv preprint arXiv:2412.12643}, 2024.

\bibitem{xu2025memory}
Mufan Xu, Gewen Liang, Kehai Chen, Wei Wang, Xun Zhou, Muyun Yang, Tiejun Zhao, and Min Zhang.
\newblock Memory-augmented query reconstruction for llm-based knowledge graph reasoning.
\newblock {\em arXiv preprint arXiv:2503.05193}, 2025.

\bibitem{yang2025qwen3}
An~Yang, Anfeng Li, Baosong Yang, Beichen Zhang, Binyuan Hui, Bo~Zheng, Bowen Yu, Chang Gao, Chengen Huang, Chenxu Lv, et~al.
\newblock Qwen3 technical report.
\newblock {\em arXiv preprint arXiv:2505.09388}, 2025.

\bibitem{yao2025learning}
Tianjun Yao, Haoxuan Li, Zhiqiang Shen, Pan Li, Tongliang Liu, and Kun Zhang.
\newblock Learning efficient and generalizable graph retriever for knowledge-graph question answering.
\newblock {\em arXiv preprint arXiv:2506.09645}, 2025.

\bibitem{yih2016value}
Wen-tau Yih, Matthew Richardson, Christopher Meek, Ming-Wei Chang, and Jina Suh.
\newblock The value of semantic parse labeling for knowledge base question answering.
\newblock In {\em Proceedings of the 54th Annual Meeting of the Association for Computational Linguistics (Volume 2: Short Papers)}, pages 201--206, 2016.

\bibitem{yu2022decaf}
Donghan Yu, Sheng Zhang, Patrick Ng, Henghui Zhu, Alexander~Hanbo Li, Jun Wang, Yiqun Hu, William Wang, Zhiguo Wang, and Bing Xiang.
\newblock Decaf: Joint decoding of answers and logical forms for question answering over knowledge bases.
\newblock {\em arXiv preprint arXiv:2210.00063}, 2022.

\bibitem{zhai2024fine}
Simon Zhai, Hao Bai, Zipeng Lin, Jiayi Pan, Peter Tong, Yifei Zhou, Alane Suhr, Saining Xie, Yann LeCun, Yi~Ma, et~al.
\newblock Fine-tuning large vision-language models as decision-making agents via reinforcement learning.
\newblock {\em Advances in neural information processing systems}, 37:110935--110971, 2024.

\bibitem{zhang2022subgraph}
Jing Zhang, Xiaokang Zhang, Jifan Yu, Jian Tang, Jie Tang, Cuiping Li, and Hong Chen.
\newblock Subgraph retrieval enhanced model for multi-hop knowledge base question answering.
\newblock {\em arXiv preprint arXiv:2202.13296}, 2022.

\bibitem{zhao2023divknowqa}
Wenting Zhao, Ye~Liu, Tong Niu, Yao Wan, Philip~S Yu, Shafiq Joty, Yingbo Zhou, and Semih Yavuz.
\newblock Divknowqa: assessing the reasoning ability of llms via open-domain question answering over knowledge base and text.
\newblock {\em arXiv preprint arXiv:2310.20170}, 2023.

\end{thebibliography}
\clearpage
\onecolumn
\appendix

\section{Algorithm}\label{sec:algorithm}

\begin{algorithm}[h]
    \caption{Dynamic MCTS-based KGQA with Path Model Pretraining and Online Refinement}
    \renewcommand{\algorithmicrequire}{\textbf{Input:}}
    \renewcommand{\algorithmicensure}{\textbf{Output:}}
    \begin{algorithmic}[1]
        \REQUIRE Question $q$, knowledge graph $\mathcal{G} = (\mathcal{E}, \mathcal{R}, \mathcal{T})$, number of selected relations $k$, MCTS iterations $N$, length of reasoning path $L$
        \ENSURE Answer set $\mathcal{A}$
        
        \STATE \texttt{/ Stage 1: Path Evaluation Model Pre-training /}
        \STATE Construct reasoning path pairs $(q, p^{+}, p^{-})$ from $\mathcal{G}$ 
        \STATE Initialize path evaluation model $S(q, \cdot; \Theta)$
        \FOR{each batch in pretraining data}
            \STATE Update $S(q, \cdot; \Theta)$ by minimizing the Pair-wise Ranking loss in Eq.(4) 
        \ENDFOR

        \STATE \texttt{/ Stage 2: Dynamic MCTS Reasoning /}
        \FOR{$i = 1$ to $N$}
            \STATE \textbf{Selection:}
            Traverse the tree from root to a leaf node by selecting child nodes according to the UCT criterion in Eq.(1)
            \STATE \textbf{Expansion:}
            
            \STATE i. At the selected node, enumerate all candidate relations from current entities and use the LLM-based planner to select the top-$k$ most relevant relations
            
            \STATE ii. Expand a new child node for each selected relation
            \STATE \textbf{Simulation:}
            For each expanded node, perform a rollout by sequentially selecting relations (guided by the path evaluation model) up to $L$ hops or until a correct answer is reached
            \STATE \textbf{Backpropagation:}
            Update the value ($w_i$) and visit ($n_i$) statistics along the traversed path from the leaf node back to the root using the score from the simulation, as per Eq.(6)
            
            \STATE \textbf{Path Evaluation Model Fine-tuning:} Generate the explored pseudo-path pairs $(\hat{p}^{+}, \hat{p}^{-})$ via Eq.(5) and fine-tune  the path evaluation model $S(q, \cdot; \Theta)$ based on Pair-wise Ranking loss in Eq.(4)
        \ENDFOR

        \STATE \texttt{/ Stage 3: Answer Extraction /}
        \STATE Collect entities reached by high-scoring reasoning paths as $\mathcal{A}$
        \STATE \textbf{return} $\mathcal{A}$
    \end{algorithmic}
    \label{algo:dynamic-mcts-final}
\end{algorithm}



\section{ Datasets}\label{sec:dataset}

\subsection{Dataset Description}
\begin{table}[h]
\centering
\label{tab:dataset_statistics}
\begin{tabular}{lccc}
\toprule
\textbf{Datasets} & \textbf{\#Train} & \textbf{\#Valid} & \textbf{\#Test} \\
\midrule
WebQSP     & 2,848   & 250    & 1,639  \\
\midrule
CWQ        & 27,639  & 3,519  & 3,531   \\
\bottomrule
\end{tabular}
\caption{Statistics of KGQA datasets.}
\label{tab:datasets}
\end{table}

We conduct extensive experiments on two widely used multi-hop Knowledge Graph Question Answering (KGQA) benchmarks: WebQSP~\cite{talmor2018web} and CWQ~\cite{yih2016value}. The statistics of these two benchmarks can be found in Table \ref{tab:datasets}, and
their details are shown as follows:

\begin{itemize}
    \item The WebQuestionsSP (WebQSP) dataset is a widely adopted benchmark for evaluating single-hop and simple multi-hop KGQA~\cite{yih2016value}. It consists of 4,837 natural language questions annotated with corresponding SPARQL queries over the Freebase knowledge graph. The dataset is partitioned into 2,848 training, 250 validation, and 1,639 test instances.
    \item The ComplexWebQuestions (CWQ) dataset is a challenging benchmark designed for multi-hop KGQA~\cite{talmor2018web}. It comprises 34,689 questions derived from WebQuestionsSP, reformulated to include more complex and compositional queries. Each question typically requires multi-step reasoning over the Freebase knowledge graph, often involving conjunctions, comparatives, or nested logical structures. The dataset is divided into 27,639 training, 3,519 validation, and 3,531 test examples.

\end{itemize}

\subsection{Data Processing}

Following prior work~\cite{shen2025reasoning,long2025enhancing, ma2025deliberation}, we preprocess the datasets by constructing localized subgraphs centered around each question entity to reduce the size of the search space. Specifically, for each question in WebQSP~\cite{yih2016value} and CWQ~\cite{talmor2018web}, we extract a subgraph from the Freebase knowledge graph by including all triples within a predefined number of hops from the topic entity. This approach preserves the essential context required for multi-hop reasoning while significantly improving computational efficiency. 

\section{ Baselines}\label{sec:baseline}

In this part, we introduce the details of the compared baselines as follows:

\begin{itemize}

\item \textbf{Semantic Parsing Methods.} We compare our \method{} with six semantic parsing methods:

\begin{itemize}
    \item \textbf{KV-Mem}: KV-Mem~\cite{miller2016key} introduce a neural architecture that stores facts as key-value pairs and enables question answering by attending over memory slots, directly retrieving relevant information to infer answers.
    \item \textbf{EmbedKGQA}: EmbedKGQA~\cite{saxena2020improving} enhances multi-hop question answering over knowledge graphs by leveraging pretrained knowledge base embeddings, enabling the model to reason over entity and relation representations without explicit path enumeration during answer prediction.
    \item \textbf{QGG}: QGG~\cite{lan2020query} generates query graphs to answer multi-hop complex questions over knowledge bases, formulating question answering as query graph prediction and enabling structured reasoning through graph matching and path ranking mechanisms.
    \item \textbf{NSM}: NSM~\cite{he2021improving} enhances multi-hop KBQA by leveraging intermediate supervision signals, decomposing questions into reasoning steps, and training a neural state machine to sequentially predict relations and entities for accurate path-based reasoning.
    \item \textbf{TransferNet}: TransferNet~\cite{shi2021transfernet} proposes a transparent framework for multi-hop QA over relational graphs by transferring question semantics to relation paths through interpretable path ranking and structured reasoning, enabling effective and explainable answer prediction.
    \item \textbf{KGT5}: KGT5~\cite{saxena2022sequence} formulates knowledge graph completion and question answering as unified sequence-to-sequence tasks, leveraging pre-trained language models to jointly encode input queries and generate answer entities or triples in a flexible and end-to-end manner.
    
\end{itemize}

\item \textbf{Retrieval-Based Methods.} We compare our \method{} with four retrieval-based methods:

\begin{itemize}
    \item \textbf{GraftNet}: GraftNet~\cite{sun2018open} proposes an early fusion framework that jointly encodes knowledge base facts and supporting text by constructing a heterogeneous graph, enabling effective reasoning through graph convolutional networks for open-domain question answering.
    \item \textbf{PullNet}: PullNet~\cite{sun2019pullnet} introduces an iterative retrieval mechanism that expands a query-specific subgraph by pulling relevant facts from both knowledge bases and text, enabling joint reasoning over heterogeneous evidence for open-domain question answering. 
    \item \textbf{SR+NSM}: SR+NSM~\cite{zhang2022subgraph} enhances multi-hop KBQA by first retrieving a question-relevant subgraph and then performing symbolic reasoning over it using Neural Symbolic Machines, improving efficiency and accuracy through constrained and focused logical inference.
    \item \textbf{SR+NSM+E2E}: SR+NSM+E2E~\cite{zhang2022subgraph} extends SR+NSM by enabling end-to-end training that jointly optimizes subgraph retrieval and reasoning. This integration enhances model coherence and allows better alignment between retrieved subgraphs and final answer prediction.
\end{itemize}

    \item  \textbf{General
Large Language Models (LLMs).} We compare our \method{} with six general
LLMs:

\begin{itemize}

    \item \textbf{Flan-T5-xl}: Flan-T5-xl~\cite{chung2024scaling} is an instruction-finetuned variant of the T5 model, trained on a diverse collection of tasks with natural language instructions. By leveraging large-scale instruction tuning, it improves zero-shot and few-shot performance across diverse NLP benchmarks.
    \item \textbf{Alpaca-7B}: Alpaca-7B~\cite{taori2023stanford} is an instruction-following language model fine-tuned from LLaMA-7B using self-instruct techniques. It demonstrates strong zero-shot and few-shot performance by aligning with human instructions across various NLP tasks.
    \item \textbf{Llama3-8B}: Llama3-8B~\cite{dubey2024llama} is part of the LLaMA 3 family of models, designed for improved instruction following, reasoning, and code generation. Pretrained on a high-quality corpus and fine-tuned with supervised signals, it achieves strong performance across diverse benchmarks.
    \item \textbf{Qwen2.5-7B}: Qwen2.5-7B~\cite{team2024qwen2} is a 7B-parameter instruction-tuned language model developed by Alibaba, optimized for tasks such as reasoning, code generation, and dialogue. It supports multi-turn conversation and demonstrates competitive performance on standard benchmarks.
    \item \textbf{ChatGPT}: ChatGPT~\cite{schulman2022chatgpt} is a conversational AI developed by OpenAI, based on the GPT architecture. It is designed to understand natural language, engage in dialogue, answer questions, and assist with a wide range of tasks across domains.
    \item \textbf{ChatGPT+CoT}: ChatGPT with Chain-of-Thought (CoT)~\cite{wei2022chain} prompting enhances the model's reasoning capabilities by encouraging it to generate intermediate reasoning steps before arriving at a final answer, improving performance on complex, multi-step problems.
\end{itemize}

    \item \textbf{LLMs with KG.} We compare our \method{} with fourteen LLMs with KG methods:

\begin{itemize}
    \item  \textbf{UniKGQA}: UniKGQA~\cite{jiang2022unikgqa} is a unified framework that integrates retrieval and reasoning for multi-hop question answering over knowledge graphs, combining subgraph retrieval, query decomposition, and neural reasoning in an end-to-end manner.
    \item \textbf{DECAF}: DECAF~\cite{yu2022decaf} is a joint framework for question answering over knowledge bases that simultaneously decodes logical forms and answers. By leveraging dual supervision, it enhances both symbolic reasoning accuracy and direct answer prediction in a unified architecture.
    \item \textbf{KD-CoT}: KD-CoT~\cite{wang2023knowledge} is a framework that enhances the faithfulness of large language models by guiding Chain-of-Thought reasoning with external knowledge, improving accuracy in knowledge-intensive question answering tasks.
    \item \textbf{Nutrea}: Nutrea~\cite{choi2023nutrea} proposes a neural tree search framework for context-guided multi-hop KGQA. It incrementally constructs reasoning trees by integrating question semantics and graph context, enabling efficient exploration of multi-hop paths for accurate answer prediction.
    \item \textbf{ToG}: ToG~\cite{sun2023think} is a framework that enables large language models to perform deep and responsible reasoning over knowledge graphs by combining structured graph information with iterative thinking and verification mechanisms for reliable multi-hop QA.
    \item \textbf{RoG}: RoG~\cite{luo2023reasoning} is a framework that enhances the faithfulness and interpretability of large language model reasoning by grounding multi-hop question answering on knowledge graphs, integrating symbolic path tracking with natural language generation.
    \item \textbf{KAPING}: KAPING~\cite{baek2023knowledge} introduces knowledge-augmented prompting by integrating structured triples into Chain-of-Thought (CoT) reasoning. It guides large language models to generate intermediate reasoning steps, enabling zero-shot multi-hop KGQA without task-specific fine-tuning.
    \item \textbf{ReasoningLM}: ReasoningLM~\cite{jiang2023reasoninglm} enhances pre-trained language models for KGQA by injecting subgraph structures into the input representation. It enables structural reasoning over retrieved subgraphs through a reasoning-aware encoder, improving performance on complex multi-hop queries.
    \item \textbf{FiDeLis}: FiDeLis~\cite{sui2024fidelis} proposes a faithfulness-aware KGQA framework that enhances reasoning consistency in LLMs by aligning generated logical forms with answer predictions. It introduces fidelity constraints to reduce hallucinations and improve answer correctness.
    
    
    \item \textbf{GNN-RAG}: GNN-RAG~\cite{mavromatis2024gnn} integrates graph neural networks with retrieval-augmented generation by encoding knowledge subgraphs into LLMs' context. It enables structural reasoning over retrieved subgraphs, improving answer accuracy in KGQA through explicit graph-aware representations.
    \item \textbf{DoG}: DoG~\cite{ma2025debate} is a flexible and reliable reasoning framework that enables large language models to generate and evaluate multiple reasoning paths over knowledge graphs through a debate-style process, enhancing robustness and answer faithfulness.
    \item \textbf{DuarL}: DuarL~\cite{liu2025dual} is a collaborative framework that integrates GNNs and LLMs for KGQA, where GNNs capture structural semantics and LLMs perform adaptive reasoning, enabling accurate and interpretable multi-hop QA.
    \item \textbf{DP}: DP~\cite{ma2025deliberation} is a trustworthy reasoning framework that guides large language models using prior knowledge from knowledge graphs. It iteratively verifies and refines reasoning paths to enhance faithfulness, robustness, and answer accuracy in KGQA.
    \item \textbf{RwT}: RwT~\cite{shen2025reasoning} is a faithful KGQA framework that models multi-hop reasoning as tree-structured exploration over knowledge graphs, enabling large language models to generate interpretable reasoning paths and improve answer consistency and accuracy.

\end{itemize}

\end{itemize}

\section{More experimental results}\label{sec:experiments_results}

To more thoroughly illustrate the impact of hyperparameter variations on model performance, we report detailed numerical results showing how performance fluctuates under different hyperparameter settings. As presented in Table~\ref{tab:sensitive_k} and Table~\ref{tab:sensitive_length}, these results provide a comprehensive understanding of the model’s sensitivity and stability across a range of configurations.

\begin{table}[t]
\centering
\small
\begin{tabular}{lcccc}
\toprule
\multirow{2}{*}{Method} & \multicolumn{2}{c}{WebQSP} & \multicolumn{2}{c}{CWQ} \\
 & Hits@1 & F1 & Hits@1 & F1 \\
\midrule
$k=2$ & 93.0 & 80.9 & 76.6 & 73.8 \\
$k=3$ & 94.0 & 81.7 & 78.0 & 75.1 \\
$k=4$ & 94.0 & 81.8 & 77.8 & 75.0 \\
$k=5$ & 93.9 & 81.7 & 78.0 & 75.2 \\
\bottomrule
\end{tabular}
\caption{Hyperparameter sensitivity analysis of the number of selected relations $k$ on the WebQSP and CWQ datasets.}
\label{tab:sensitive_k}
\end{table}

\begin{table}[t]
\centering
\small
\begin{tabular}{lcccc}
\toprule
\multirow{2}{*}{Method} & \multicolumn{2}{c}{WebQSP} & \multicolumn{2}{c}{CWQ} \\
 & Hits@1 & F1 & Hits@1 & F1 \\
\midrule
$L=2$ & 93.6 & 81.2 & 77.4 & 74.5 \\
$L=3$ & 94.0 & 81.7 & 77.6 & 74.7 \\
$L=4$ & 93.7 & 81.4 & 78.0 & 75.1 \\
$L=5$ & 93.8 & 81.6 & 77.9 & 74.9 \\
\bottomrule
\end{tabular}
\caption{Hyperparameter sensitivity analysis of the reasoning path length $L$ on the WebQSP and CWQ datasets.}
\label{tab:sensitive_length}
\end{table}

\section{Prompt Template}

We provide the prompt templates used by the LLM-based planner to select the top-$k$ most relevant relations from the candidate set at each step of path expansion in Fig. 3, as part of the LLM Guided Path Expansion module.

\begin{tcolorbox}[title=Prompt Template for LLM-Guided Path Expansion, colback=white, colframe=black!75, boxrule=0.5pt, arc=2mm]

\textbf{Role} \\
You are an expert assistant for Knowledge Graph Question Answering (KGQA). Your core capability is to deeply understand natural language questions and the semantics of knowledge graph relations to find the most relevant reasoning paths.

\textbf{Task} \\
Your task is to act as a \textbf{“Relation Retriever.”} Given a natural language question and a list of candidate relations, you must analyze the semantics of the question and each relation to select up to \texttt{k} relations that are most likely to lead to the correct answer.

\textbf{Rules and Constraints}
\begin{itemize}[leftmargin=1.5em, itemsep=0pt, topsep=0pt]
    \item \textbf{Fidelity to Candidates}: Your selection of relations \textbf{MUST} come strictly from the provided \texttt{Candidate Relations} list. Do not invent or modify relations.
    \item \textbf{Quantity Limit}: Return no more than \texttt{k} relations. If multiple relations are highly relevant, order them from most to least relevant. If there are fewer than \texttt{k} relevant relations, return only those.
    \item \textbf{Output Format}: Your response \textbf{MUST} be a list of strings, containing the names of the relations you have selected.
\end{itemize}

\textbf{Example}
\begin{itemize}[leftmargin=1.5em, itemsep=0pt, topsep=2pt]
    \item \textbf{Input:}
    \begin{itemize}
        \item \texttt{Question}: "who was the president after jfk died"
        \item \texttt{Candidate Relations}: \{"government.president", "government.president.successor", "location.location.containedby", "people.person.place\_of\_birth"\}
        \item \texttt{K}: 2
    \end{itemize}
    
    \item \textbf{Output:}
\begin{verbatim}
["government.president", "government.president.successor"]
\end{verbatim}
\end{itemize}

\textbf{Your Task}
\begin{itemize}[leftmargin=1.5em, itemsep=0pt, topsep=2pt]
    \item \texttt{Question}: \{question\}
    \item \texttt{Candidate Relations}: \{relations\_list\}
    \item \texttt{K}: \{k\}
\end{itemize}

\textbf{Output:}
\begin{verbatim}
[]
\end{verbatim}
\end{tcolorbox}
\captionof{figure}{Prompt template used in the LLM-based planner to select top-k relations during reasoning.}
\label{fig:prompt_llm_planner}

\section{The Use of Large Language Models (LLMs)}

Large language models (LLMs) were only used to improve the clarity, grammar, and fluency of the manuscript. They were not involved in the development of research ideas, experimental design, data analysis, or any other aspect of the scientific content.
\end{document}